\documentclass[conference]{IEEEtran}
\IEEEoverridecommandlockouts

\usepackage{cite}
\usepackage{amsmath,amssymb,amsfonts}
\usepackage{algorithmic}
\usepackage{graphicx}
\usepackage{textcomp}
\usepackage{xcolor}
\usepackage{ragged2e}
\usepackage{multirow}
\def\BibTeX{{\rm B\kern-.05em{\sc i\kern-.025em b}\kern-.08em
    T\kern-.1667em\lower.7ex\hbox{E}\kern-.125emX}}
\newcommand{\ts}{\textsuperscript}
\begin{document}

\title{BIORTHOGONAL TUNABLE WAVELET UNIT WITH LIFTING SCHEME IN CONVOLUTIONAL NEURAL NETWORK
\thanks{This work was supported by Innovative Human Resource Development for Local Intellectualization program through the Institute of Information \& Communications Technology Planning \& Evaluation (IITP) grant funded by the Korea government (MSIT)(IITP-2024-2020-0-01741).}
}

\author{\IEEEauthorblockN{An Le\ts{1}, Hung Nguyen\ts{1}, Sungbal Seo\ts{2}, You-Suk Bae\ts{2}, Truong Nguyen\ts{1}}
\IEEEauthorblockA{\ts{1}Electrical and Computer Engineering Department, University of California San Diego, La Jolla, CA 92093, USA \\
\{d0le,hun004,tqn001\}@ucsd.edu\\
\ts{2}Department of Computer Engineering, Tech University of Korea, Siheung 15073, Korea\\
\{sungbal,ysbae\}@tukorea.ac.kr}
}
\maketitle

\begin{abstract}
This work introduces a novel biorthogonal tunable wavelet unit constructed using a lifting scheme that relaxes both the orthogonality and equal filter length constraints, providing greater flexibility in filter design. The proposed unit enhances convolution, pooling, and downsampling operations, leading to improved image classification and anomaly detection in convolutional neural networks (CNN). When integrated into an 18-layer residual neural network (ResNet-18), the approach improved classification accuracy on CIFAR-10 by 2.12\% and on the Describable Textures Dataset (DTD) by 9.73\%, demonstrating its effectiveness in capturing fine-grained details. Similar improvements were observed in ResNet-34. For anomaly detection in the hazelnut category of the MVTec Anomaly Detection dataset, the proposed method achieved competitive and well-balanced performance in both segmentation and detection tasks, outperforming existing approaches in terms of accuracy and robustness.
\end{abstract}
\section{Introduction}
\label{sec:intro}
\begin{figure}[t!]
\begin{center}
\includegraphics[width=\linewidth]{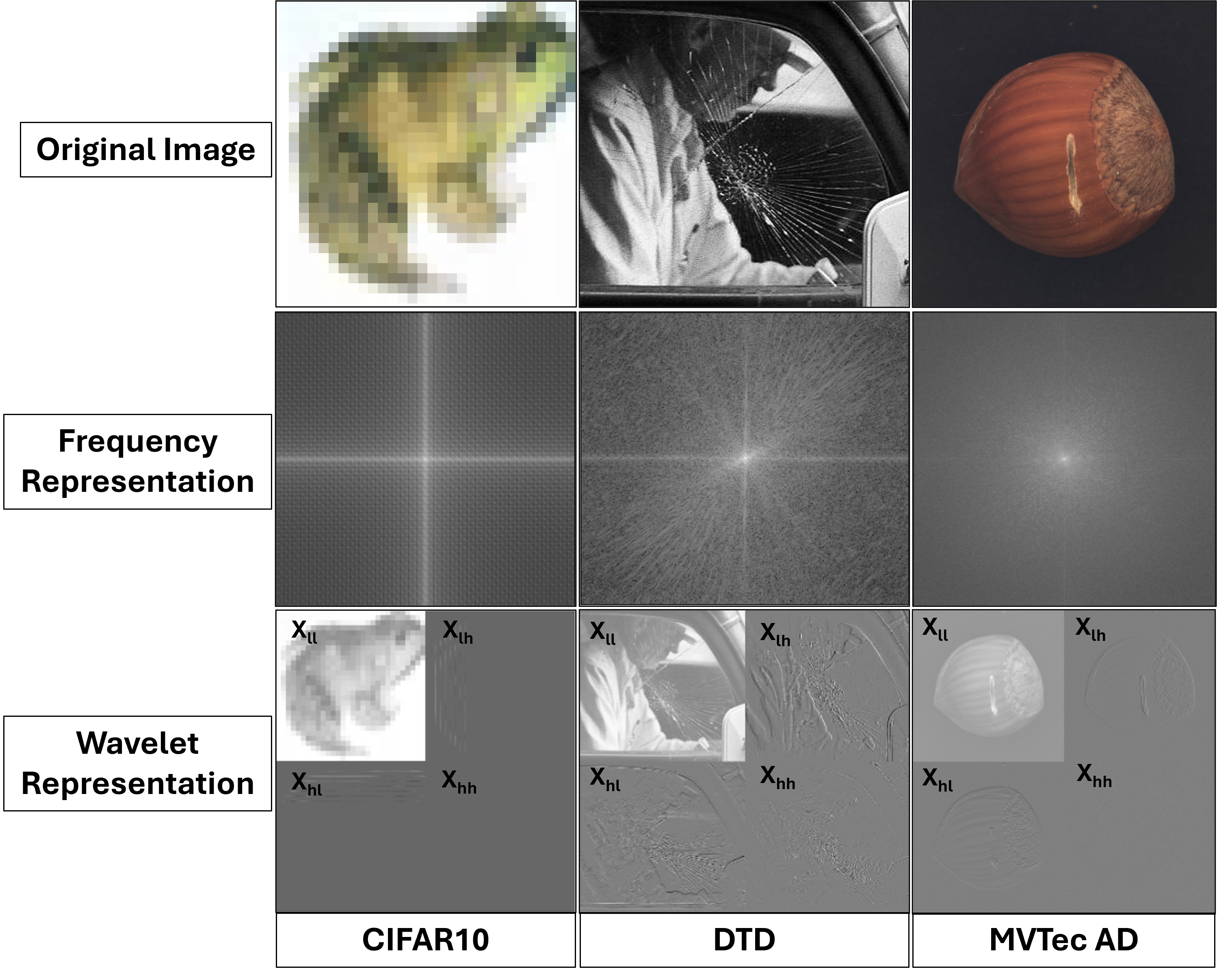}
\end{center}
\caption{From left to right, wavelet (Haar) and frequency representations of the samples from CIFAR10 (first column), DTD (second column), and MVTecAD (third column). The original images (top row) are shown with its frequency representation (middle row) and wavelet representation (bottom row). $\textbf{X}_{ll}$, $\textbf{X}_{lh}$, $\textbf{X}_{hl}$, and $\textbf{X}_{hh}$ show the coarse approximation and details wavelet representations.} 
\label{fig:Frequency_Wavelet_Analysis}
\end{figure}
Max pooling, a key component in CNN architectures such as ResNets \cite{Resnet}, emphasizes dominant features but discards fine details, leading to aliasing artifacts \cite{Alias_CNN}. While frequency-based methods \cite{Spectral_Pooling, diffstride} focus on low-frequency components, wavelet-based models like WaveCNet \cite{WaveCNET} predominantly use low-pass filters. However, models such as Wavelet-Attention CNNs \cite{Wavelet-AttentionCNN} incorporates both coarse and fine-grained details, which is crucial for high-resolution image processing.

As depicted in Fig. \ref{fig:Frequency_Wavelet_Analysis}, the CIFAR-10 dataset \cite{CIFAR10} predominantly consists of low-frequency information, whereas MVTecAD \cite{MVTecAD_1,MVTecAD_2} and DTD \cite{DTD} exhibit features distributed across both low- and high-frequency domains. In the "cracked" DTD sample shown in the second column of Fig. \ref{fig:Frequency_Wavelet_Analysis}, the low-pass component $\textbf{X}_{ll}$ retains only minimal texture details, whereas the high-pass components $\textbf{X}_{hl}$, $\textbf{X}_{lh}$, and $\textbf{X}_{hh}$ effectively capture its distinctive features. This underscores the importance of maintaining both high- and low-frequency information within CNN architectures.
\begin{figure}[b]
\begin{center}
\includegraphics[width=\linewidth]{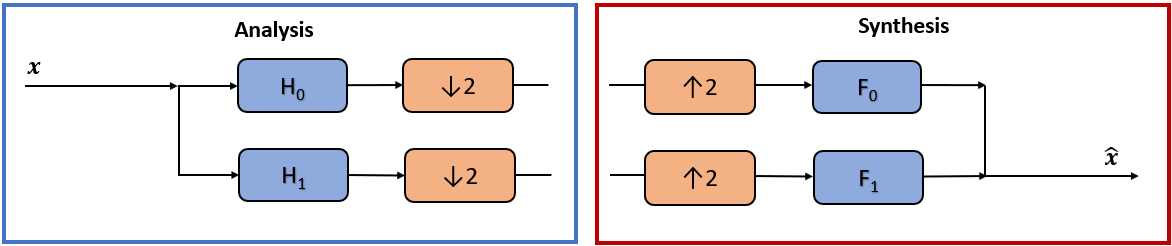}
\end{center}
\caption{Two-channel filter bank architecture.}
\label{fig:filterbank}
\end{figure}
\begin{figure}[t!]
\begin{center}
\includegraphics[width=\linewidth]{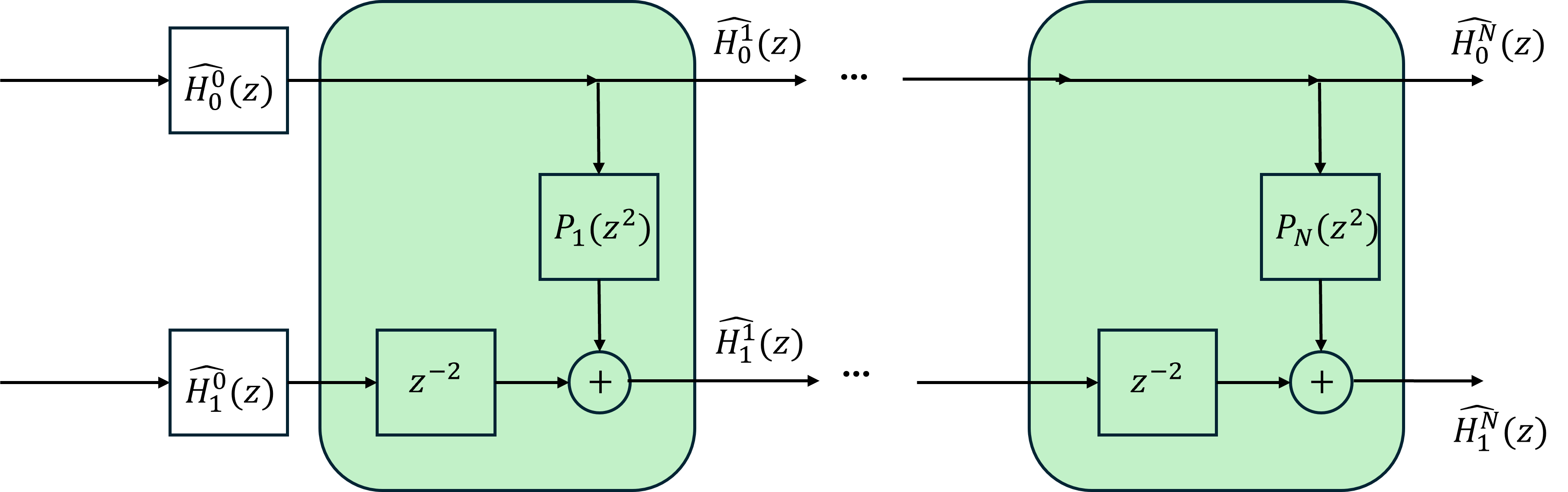}
\end{center}
\caption{The analysis of a biorthogonal filter bank constructed with lifting scheme. $P_k(z)$ is the lifting step function for k in the range from 1 to $N$.}
\label{fig:LiftingScheme}
\end{figure}
Previous studies \cite{AnLe, OrthLatt_UwU} used wavelet decomposition and perfect reconstruction to retain full image information, improving performance. While tunable wavelet filters \cite{AnLe, OrthLatt_UwU} enhanced CNNs, especially for high-frequency images, they rely on orthogonal wavelets and require equal filter lengths. To overcome these limitations, we propose a biorthogonal tunable wavelet unit based on a lifting scheme (LS-BiorUwU), which relaxes these constraints, allowing greater design flexibility. Integrated into ResNet architectures, the proposed unit improves classification on CIFAR-10 and DTD and serves as a feature extractor in the CFLOW-AD anomaly detection pipeline \cite{CFLOW_Gudovskiy_2022_WACV}, tested on the hazelnut category of MVTecAD \cite{MVTecAD_1,MVTecAD_2}. Our approach enhances CNN performance in both image classification and anomaly detection. In summary:
\begin{itemize}
    \item We propose LS-BiorUwU, a novel biorthogonal tunable wavelet unit based on the lifting scheme that relaxes the orthogonality and equal filter length constraints imposed by existing orthogonal wavelet units.
    \item We integrate the proposed unit into ResNet architectures trained on CIFAR-10 and DTD, achieving strong classification performance, especially on DTD.
    \item We incorporate the unit into the CFLOW-AD anomaly detection model and evaluate it on the MVTecAD dataset.
\end{itemize}
\begin{figure}[!b]
\begin{center}
\includegraphics[width=\linewidth]{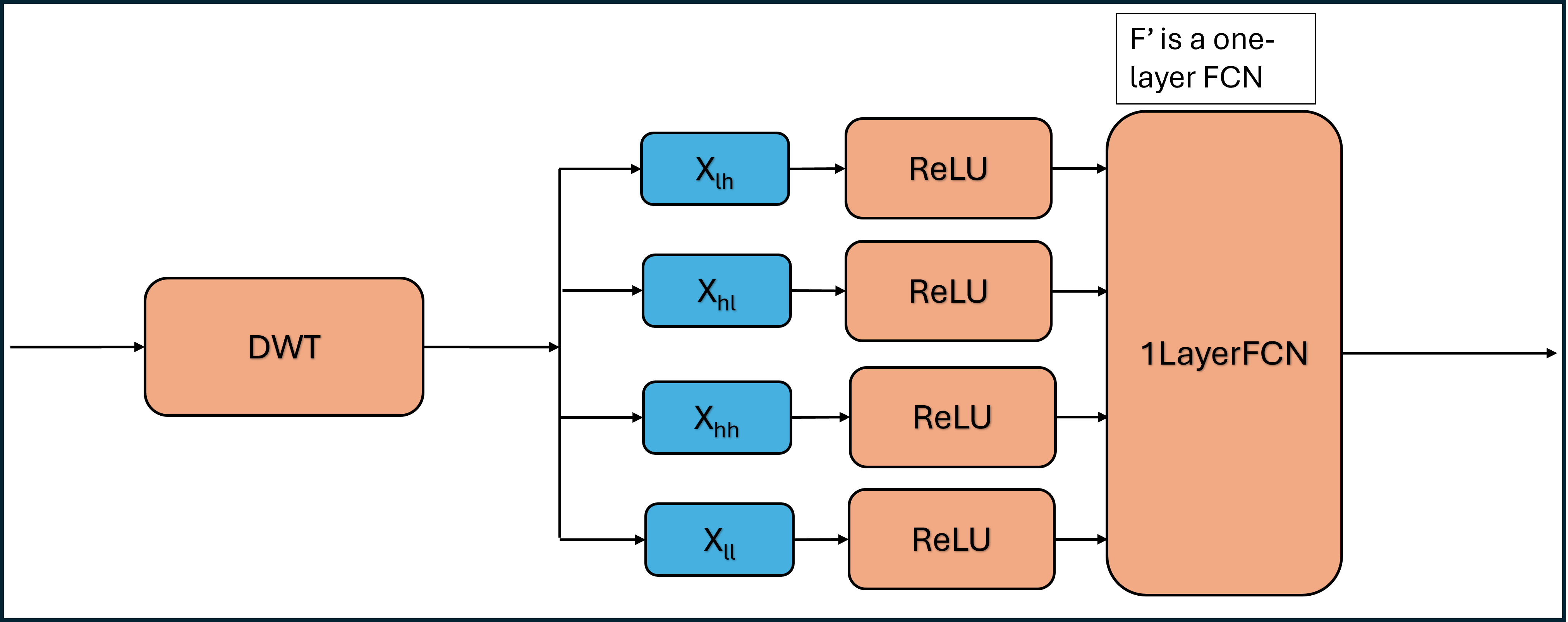}
\end{center}
\caption{\justifying{Diagram of low-pass and high-pass component implementation. The signal goes from left to right. The results from DWT go to ReLU functions to become the inputs of a one-layer FCN. Because an FCN can take inputs of arbitrary sizes, the one-layer FCN can read the decomposed components and finetune the trainable coefficients to optimally combine the decomposed components. The fine-tuned one-layer FCN combines the inputs to find the optimal feature map.}}
\label{fig:OneLayerFCN}
\end{figure}
\section{Related Works}
\label{sec:relatedwork}
Max pooling in CNNs downsamples feature maps by selecting maximum values, preserving key features \cite{Feature-Pooling_in_Visual-Recognition, MaxPool}. However, without filtering, it introduces aliasing artifacts, leading to frequency overlap and Moiré patterns \cite{Alias_CNN}, and can distort object structures in deeper networks \cite{WaveCNET}. Wavelet-based methods apply discrete wavelet transforms (DWT, FWT) to process features in the wavelet domain \cite{Mallat,Wavelets_and_filter_banks}, improving image classification \cite{WaveCNET,Wavelet-AttentionCNN,wavelet_rp}. Existing approaches rely on predefined wavelet functions, primarily using approximation components \cite{WaveCNET}, with limited reconstruction from higher-order decompositions. Wavelet-Attention CNN \cite{Wavelet-AttentionCNN} incorporates attention maps from detail components. Trainable wavelet filters, such as those in \cite{AnLe}, relaxed perfect reconstruction constraints, while the units in \cite{OrthLatt_UwU} enforced an orthogonal lattice structure. However, these methods rely on orthogonal wavelets and require equal filter lengths.

To address these limitations, we propose a biorthogonal tunable wavelet unit based on a lifting scheme, which removes these constraints and enables greater flexibility in wavelet design.

\section{Proposed Method}
\begin{figure}[!t]
\begin{center}
\includegraphics[width=\linewidth]{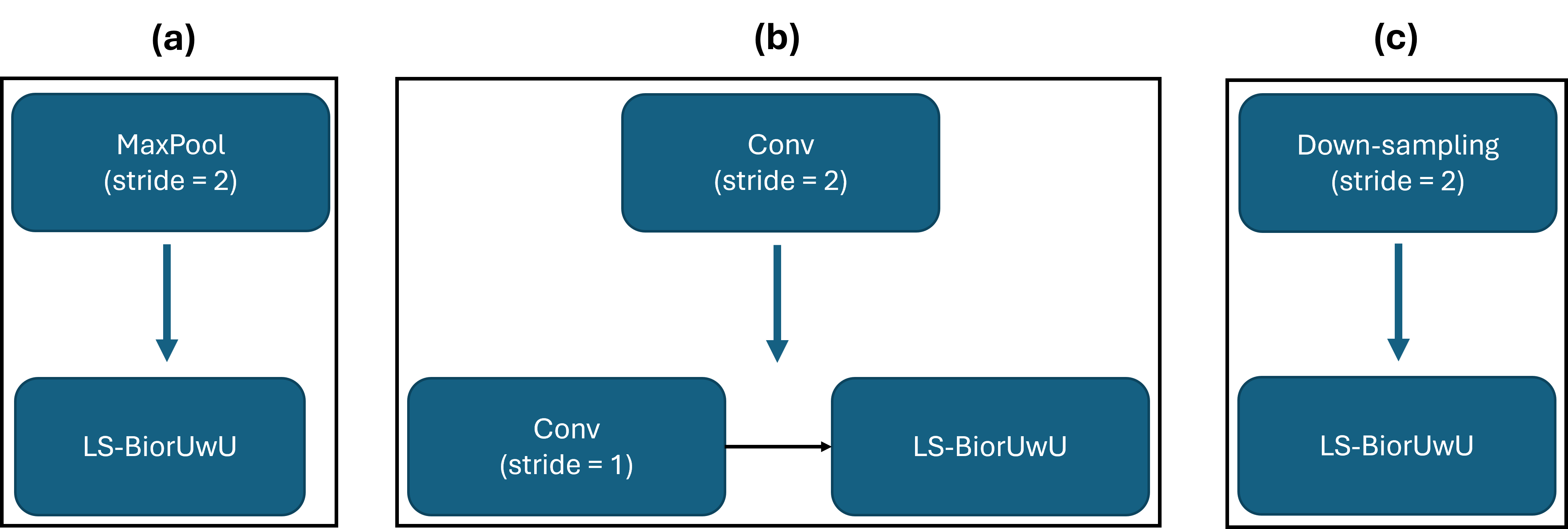}
\end{center}
\caption{Implementation of the proposed unit in CNN architecture, replacing max-pool (a), stride-convolution (b), and downsampling (c) functions.}
\label{fig:UwU2}
\end{figure}
\label{sec:method}
\subsection{Lifting Scheme for Tunable Biorthogonal Wavelet Filters}
\begin{table*}[!t]
\caption{\justifying{The recursive implementation of the lifting scheme for 1 and 2 lifting steps is tested by comparing its $\textbf{h}_0$ and $\textbf{h}_1$ coefficients with those of Bior1.3 and Bior1.5.
}}
\centering
\scalebox{1}{
\begin{tabular}{|c|c|c|c|}
\hline
\multirow{4}{*}{\textbf{1 Lifting Step}} 
& \multirow{2}{*}{\textbf{Recursive Lifting Scheme}} & \multicolumn{1}{|c|}{$\textbf{h}_0$} & \multicolumn{1}{|c|}{0.7071, 0.7071} \\\cline{3-4}
& \multicolumn{1}{}{}& \multicolumn{1}{|c|}{\textbf{$\textbf{h}_1$}} & \multicolumn{1}{|c|}{-0.0880, -0.0880, 0.7071, -0.7071, 0.0880, 0.0880}\\\cline{2-4}

& \multirow{2}{*}{\textbf{Bior1.3 Ground-Truth}}  & \multicolumn{1}{|c|}{$\textbf{h}_0$} & \multicolumn{1}{|c|}{0.7071, 0.7071} \\\cline{3-4}
& \multicolumn{1}{}{}& \multicolumn{1}{|c|}{$\textbf{h}_1$} & \multicolumn{1}{|c|}{-0.0880, -0.0880, 0.7071, -0.7071, 0.0880, 0.0880}\\\hline

\multirow{4}{*}{\textbf{2 Lifting Steps}} 
& \multirow{2}{*}{\textbf{Recursive Lifting Scheme}} & \multicolumn{1}{|c|}{$\textbf{h}_0$} & \multicolumn{1}{|c|}{0.7071, 0.7071} \\\cline{3-4}
& \multicolumn{1}{}{}& \multicolumn{1}{|c|}{\textbf{$\textbf{h}_1$}} & \multicolumn{1}{|c|}{0.0166, 0.0166, -0.1215, -0.1215, 0.7071, -0.7071, 0.1215, 0.1215, -0.0166, -0.0166
}\\\cline{2-4}

& \multirow{2}{*}{\textbf{Bior1.5 Ground-Truth}}  & \multicolumn{1}{|c|}{$\textbf{h}_0$} & \multicolumn{1}{|c|}{0.7071, 0.7071} \\\cline{3-4}
& \multicolumn{1}{}{}& \multicolumn{1}{|c|}{$\textbf{h}_1$} & \multicolumn{1}{|c|}{0.0166, 0.01666, -0.12156, -0.12156, 0.70716, -0.70716, 0.12156, 0.12156, -0.01666, -0.0166
}\\\hline

\end{tabular}}
\label{table:InitialCoef}
\end{table*}
Using a lifting scheme, the tunable biorthogonal wavelet unit (LS-BiorUwU) relaxes the constraint of orthogonality to biorthogonality and allows unequal filter lengths in the filter bank of the wavelet unit.

In the filter bank structure demonstrated in Fig. \ref{fig:filterbank}, the analysis, shown in the blue rectangle box, and synthesis, shown in the red rectangle box, parts of the filter bank have the function of decomposing and reconstructing signals, respectively. $H_{0}$ and $H_{1}$ are, correspondingly, low-pass and high-pass filters for the analysis part of the filter bank; whereas $F_{0}$ and $F_{1}$ are, respectively, low-pass and high-pass filters for the synthesis part of the filter bank. With $L$ taps, $H_{0}$ and $H_{1}$ have $\textbf{h}_{0}=[h_0(0), h_0(1),..., h_0(L-1)]$ and $\textbf{h}_{1}=[h_1(0), h_1(1),..., h_1(L-1)]$ as their coefficients, respectively. In orthogonal wavelets, $\textbf{h}_{0}$ and $\textbf{h}_{1}$ are required to have the same length and are related. This requirement can be relaxed if the filter bank is constructed with a lifting scheme from an orthogonal wavelet. Hence, $H_{0}$ and $H_{1}$ can be represented as a matrix multiplication as follows:
\begin{align}
\label{eq:LiftingScheme}
 \begin{bmatrix}
   {H}_{0}(z) \\
   {H}_{1}(z) \\
   \end{bmatrix} = \begin{bmatrix}
   \widehat{{H}^{N}_{0}}(z) \\
   \widehat{{H}^{N}_{1}}(z) \\
   \end{bmatrix}= \begin{bmatrix} 1&0 \\P_N(z^2)&1 \\ \end{bmatrix} \begin{bmatrix} 1&0 \\0&z^{-2} \\ \end{bmatrix} \cdots \nonumber\\
  \begin{bmatrix} 1&0 \\P_1(z^2)&1 \\ \end{bmatrix} \begin{bmatrix} 1&0 \\0&z^{-2} \\ \end{bmatrix}
  \begin{bmatrix}
   \widehat{{H}^{0}_{0}}(z) \\
   \widehat{{H}^{0}_{1}}(z) \\
 \end{bmatrix},
\end{align}
in which $\widehat{{H}^{N}_{0}}(z)$ and $\widehat{{H}^{N}_{1}}(z)$ are the final filter pairs constructed after $N$ lifting steps from the $\widehat{{H}^{0}_{0}}(z)$ and $\widehat{{H}^{0}_{1}}(z)$ pair of an orthogonal wavelet. In addition, for $k$ in the range from 1 to $N$, $P_k(z)$ is the lifting step function, which can be represented as follows:
\begin{equation}
\label{eq:LiftingStep}
P_k(z) = -a_k + a_kz^{-2k}\mbox{ for }k \mbox{ in }[1, N],
\end{equation}
where $a_k$ is the tunable parameter in the biorthogonal wavelet unit. The lifting scheme can be visualized in Fig. \ref{fig:LiftingScheme}. In addition, the proposed tunable biorthgonal wavelet with lifting scheme can be implemented as the following recursive algorithm:
\begin{align}
\label{eq:recusrive}
&\begin{bmatrix}
   \widehat{{H}^{k}_{0}}(z) \\
   \widehat{{H}^{k}_{1}}(z) \\
   \end{bmatrix}= \begin{bmatrix} 1&0 \\P_k(z^2)&1 \\ \end{bmatrix} \begin{bmatrix} 1&0 \\0&z^{-2} \\ \end{bmatrix}\begin{bmatrix}
   \widehat{{H}^{k-1}_{0}}(z) \\
   \widehat{{H}^{k-1}_{1}}(z) \\
   \end{bmatrix}\nonumber\\
   &=\begin{bmatrix}
   \widehat{{H}^{k-1}_{0}}(z) \\
   -a_k\widehat{{H}^{k-1}_{0}}(z)+z^{-2}\widehat{{H}^{k-1}_{1}}(z)+a_kz^{-4k}\widehat{{H}^{k-1}_{0}}(z) \\
   \end{bmatrix},
\end{align}
for $k$ in the range from 1 to $N$. In this work, Haar or Bior1.1 is used for $\widehat{{H}^{0}_{0}}(z)$ and $\widehat{{H}^{0}_{1}}(z)$ initialization. 

\subsection{2D Implementation}

From the coefficients \( \textbf{h}_0 \) and \( \textbf{h}_1 \), the high-pass and low-pass filter matrices \( \textbf{H} \) and \( \textbf{L} \) are derived to compute the approximation component \( \textbf{X}_{ll} \) and the detail components \( \textbf{X}_{lh} \), \( \textbf{X}_{hl} \), and \( \textbf{X}_{hh} \) of the input \( \textbf{X}\). The matrix \( \textbf{L} \) is computed as follows:
\begin{equation}
\label{eq:L}
 \textbf{L} = \textbf{D}\widehat{\textbf{H}},
\end{equation}
where \( \textbf{D} \) denotes the downsampling matrix, and \( \widehat{\textbf{H}} \)  is a Toeplitz matrix formed using the filter coefficients of \( \textbf{H}_{0}(z) \). The matrix \( \textbf{H} \) follows the same structure as \( \textbf{L} \) but is derived from the filter coefficients of \( \textbf{H}_{0}(z^{-1}) \). Using \( \textbf{H} \) and \( \textbf{L} \), the components \( \textbf{X}_{ll} \), \( \textbf{X}_{lh} \), \( \textbf{X}_{hl} \), and \( \textbf{X}_{hh} \) are computed as follows:
\begin{equation}
\label{eq:decomposition}
\begin{aligned}
 \textbf{X}_{ll} &=  \textbf{L}\textbf{X}\textbf{L}^T, &
 \textbf{X}_{lh} &=  \textbf{H}\textbf{X}\textbf{L}^T,\\
 \textbf{X}_{hl} &=  \textbf{L}\textbf{X}\textbf{H}^T, &
 \textbf{X}_{hh} &=  \textbf{H}\textbf{X}\textbf{H}^T.
\end{aligned}
\end{equation}

\subsection{Implementation in CNN architectures}
The proposed units were incorporated into ResNet family architectures. In downsampling and pooling layers, the UwU is followed by a one-layer fully connected network (FCN), as depicted in Fig. \ref{fig:OneLayerFCN}. Furthermore, the stride-2 convolution is replaced with a non-stride convolution block, followed by the proposed LS-BiorUwU, as shown in Fig. \ref{fig:UwU2}.

\section{Experiments and Results}
\label{sec:exandresults}

This section integrates the proposed LS-BiorUwU unit into CNN architectures. First, the lifting scheme implementation for initialization is examined, followed by an analysis of the frequency response of the tuned coefficients. Next, the unit is applied to ResNet18 (tested on CIFAR-10 and DTD with 1, 2, and 3 lifting steps) and ResNet34 (with 1, 2, and 3 lifting steps). Finally, its effectiveness is evaluated within the CFLOW-AD pipeline for anomaly detection on the Hazelnut class in the MVTecAD dataset.
\subsection{Lifting Scheme for Tunable Biorthogonal Wavelet Filters Coefficient Analysis}

\begin{figure*}[!t]
\begin{center}
\includegraphics[width=1\linewidth]{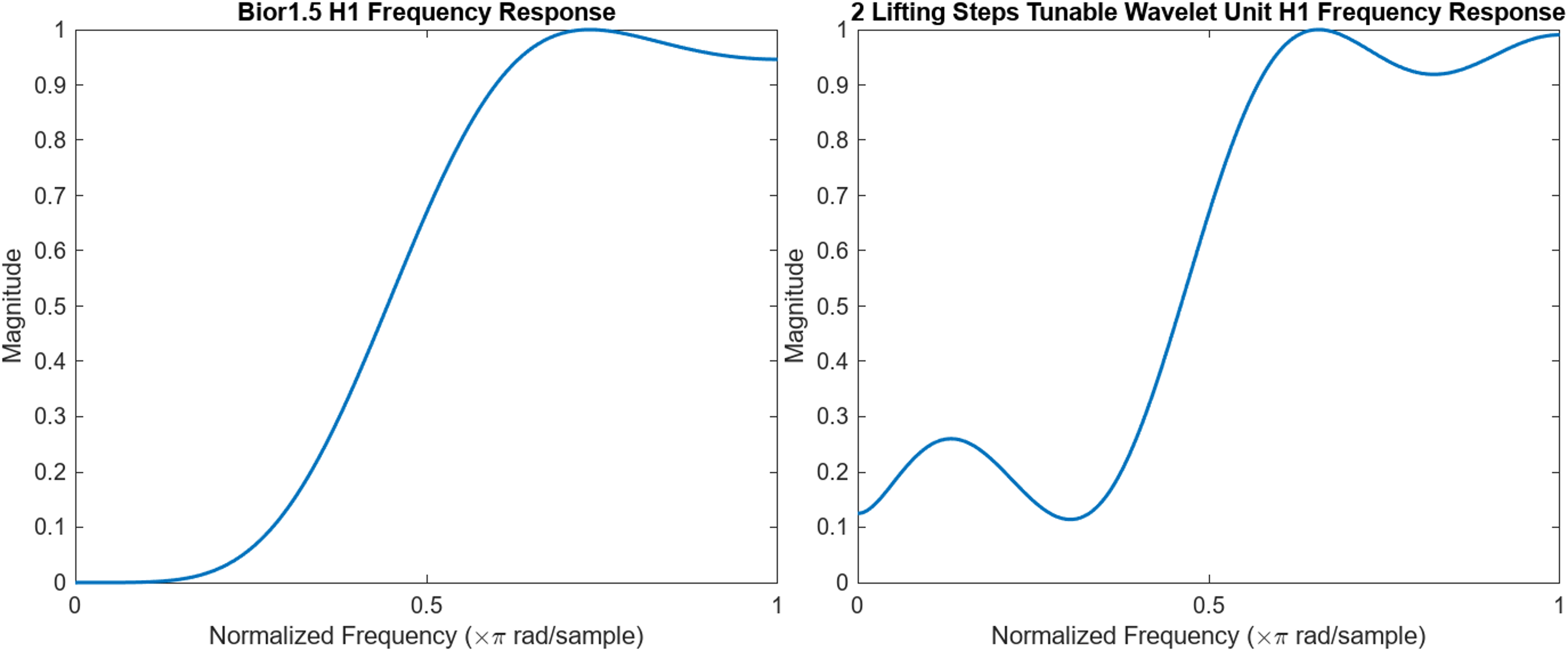}
\end{center}
\caption{Frequency Response Analysis of the Bior1.5 $H_1$ (left) and the tuned coefficients of $H_1$ of LS-BiorUwU after 2 lifting steps (right).} 
\label{fig:Frequency_Response_Analysis}
\end{figure*}
This section examines the recursive implementation of the lifting scheme for LS-BiorUwU, starting from the Haar/Bior1.1 wavelet with 1 and 2 lifting steps. Higher lifting steps are not considered due to diminishing returns in classification improvement. The coefficients used are $a_1 = \frac{880}{7071}$ for 1 step and $a_1 = \frac{405}{2357}$, $a_2 = \frac{-166}{7071}$ for 2 steps. Table \ref{table:InitialCoef} shows that the resulting $\textbf{h}_0$ and $\textbf{h}_1$ closely match Bior1.3 and Bior1.5, confirming the proposed scheme's ability to construct biorthogonal wavelets with correct lifting steps and initial orthogonal wavelet filters. The LS-BiorUwU unit is then integrated into a ResNet18 pooling layer trained on CIFAR-10, and its 2-step version (LS-BiorUwU-2Step) is analyzed. As shown in Fig. \ref{fig:Frequency_Response_Analysis}, the tuned $H_1$ filter retains high-pass characteristics.
\subsection{Image Classification: CIFAR10 and DTD}
This section evaluates the performance of LS-BiorUwU units with 1, 2, and 3 lifting steps in ResNet architectures on CIFAR-10 and DTD. CIFAR-10 \cite{CIFAR10} consists of 60,000 low-resolution (32×32) color images across 10 classes, with 50,000 for training and 10,000 for testing. In contrast, DTD \cite{DTD} (Describable Textures Dataset) includes 5,640 high-resolution images across 47 categories, emphasizing rich textures and high-frequency details. The LS-BiorUwU units were primarily tested on ResNet18 and compared to the tunable orthogonal lattice wavelet unit (OrthLatt-UwU) with 2-tap filters initialized with Haar/Bior1.1 \cite{OrthLatt_UwU}. For LS-BiorUwU-1Step and LS-BiorUwU-2Step, Bior1.3 and Bior1.5 filter coefficients were used to determine $a_1$ and $a_2$ for initialization. For LS-BiorUwU-3Step, the $a_1$ and $a_2$ values from LS-BiorUwU-2Step were used, while $a_3$ was set to a value near zero for initialization.
\begin{table}[t!]
\caption{\justifying{Accuracy of LS-BiorUwU with 1, 2 and 3 lifting steps, along with OrthLatt-UwU-2Tap evaluated on ResNet18 for DTD and CIFAR10.}}
\centering
\scalebox{0.8}{
\begin{tabular}{cccc}
\hline
\multicolumn{4}{|c|}{\textbf{DTD}} \\ \hline
\multicolumn{4}{|c|}{Baseline: 33.99$\%$} \\ \hline
\multicolumn{1}{|c|}{OrthLatt-UwU-2Tap} & \multicolumn{1}{|c|}{LS-BiorUwU-1Step} & \multicolumn{1}{|c|}{LS-BiorUwU-2Step} & \multicolumn{1}{|c|}{LS-BiorUwU-3Step}\\ \hline
\multicolumn{1}{|c|}{40.37$\%$} & \multicolumn{1}{|c|}{42.71$\%$} & \multicolumn{1}{|c|}{43.67$\%$}& \multicolumn{1}{|c|}{43.72$\%$} \\ \hline\hline
\multicolumn{4}{|c|}{\textbf{CIFAR10}}\\ \hline
\multicolumn{4}{|c|}{Baseline: 92.44$\%$}\\ \hline
\multicolumn{1}{|c|}{OrthLatt-UwU-2Tap} & \multicolumn{1}{|c|}{LS-BiorUwU-1Step} & \multicolumn{1}{|c|}{LS-BiorUwU-2Step} & \multicolumn{1}{|c|}{LS-BiorUwU-3Step}\\ \hline
\multicolumn{1}{|c|}{94.97$\%$} & \multicolumn{1}{|c|}{94.08$\%$} & \multicolumn{1}{|c|}{94.56$\%$}& \multicolumn{1}{|c|}{94.39$\%$} \\ \hline
\end{tabular}}
\label{table:Cifar10_DTD}
\end{table}
\subsubsection{ResNet18}

LS-BiorUwU with 1, 2, and 3 lifting steps was implemented in ResNet18 and tested on CIFAR-10 and DTD, representing datasets with low-resolution, low-pass features and high-resolution, high-pass details, respectively. As shown in Table \ref{table:Cifar10_DTD}, LS-BiorUwU-ResNet18 outperformed the baseline ResNet18 across all lifting step initializations. While LS-BiorUwU did not surpass OrthLatt-UwU-2Tap on CIFAR-10, it showed a clear advantage on DTD. Nevertheless, it shows that the proposed LS-BiorUwU can still perform well on low-resolution data.

For DTD, which contains high-resolution images with detailed features, increasing the lifting steps and the order of the high-pass filter led to better results. Since OrthLatt-UwU-2Tap (Bior1.1/Haar) and LS-BiorUwU (1, 2, and 3 lifting steps) share the same 2-Tap low-pass filter length, adding more lifting steps improved DTD accuracy by increasing the order of the high-pass filter. However, LS-BiorUwU-3Step provided only limited improvement over LS-BiorUwU-2Step, likely due to suboptimal initialization in LS-BiorUwU-3Step. These results suggest that LS-BiorUwU maintains competitive performance on low-resolution datasets, while demonstrating superior accuracy on high-resolution images that contain rich details and high-frequency features—benefiting from the use of higher-order high-pass filters.

\subsubsection{Extended Study with ResNet34}
\begin{table}[t!]
\caption{\justifying{Accuracy of LS-BiorUwU with 1, 2 and 3 lifting steps, along with OrthLatt-UwU-2Tap evaluated on ResNet34 for DTD and CIFAR10.}}
\centering
\scalebox{0.8}{
\begin{tabular}{cccc}
\hline
\multicolumn{4}{|c|}{\textbf{DTD}} \\ \hline
\multicolumn{4}{|c|}{Baseline: 24.47$\%$} \\ \hline
\multicolumn{1}{|c|}{OrthLatt-UwU-2Tap} & \multicolumn{1}{|c|}{LS-BiorUwU-1Step} & \multicolumn{1}{|c|}{LS-BiorUwU-2Step} & \multicolumn{1}{|c|}{LS-BiorUwU-3Step}\\ \hline
\multicolumn{1}{|c|}{41.49$\%$} & \multicolumn{1}{|c|}{41.76$\%$} & \multicolumn{1}{|c|}{41.76$\%$}& \multicolumn{1}{|c|}{42.45$\%$} \\ \hline\hline
\multicolumn{4}{|c|}{\textbf{CIFAR10}}\\ \hline
\multicolumn{4}{|c|}{Baseline: 94.33$\%$}\\ \hline
\multicolumn{1}{|c|}{OrthLatt-UwU-2Tap} & \multicolumn{1}{|c|}{LS-BiorUwU-1Step} & \multicolumn{1}{|c|}{LS-BiorUwU-2Step} & \multicolumn{1}{|c|}{LS-BiorUwU-3Step}\\ \hline
\multicolumn{1}{|c|}{95.44$\%$} & \multicolumn{1}{|c|}{94.45$\%$} & \multicolumn{1}{|c|}{94.56$\%$}& \multicolumn{1}{|c|}{94.36$\%$} \\ \hline
\end{tabular}}
\label{table:DTD_CIFAR10_ResNet34}
\end{table}
In this section, LS-BiorUwU with 1, 2, and 3 lifting steps was implemented in ResNet34, tested on CIFAR-10 and DTD, and compared against the baseline ResNet34 and OrthLatt-UwU-2Tap. As shown in Table \ref{table:DTD_CIFAR10_ResNet34}, while LS-BiorUwU achieves comparable performance on CIFAR-10, it consistently outperforms both baseline ResNet34 and OrthLatt-UwU-2Tap ResNet34 at every lifting step on DTD. Additionally, increasing the number of lifting steps leads to better performance, aligning with the trends observed in the ResNet18 experiments. This suggests that the performance gains of LS-BiorUwU extend to deeper neural network architectures as well.
\begin{table}[b!]
\caption{\justifying{Segmentation and Detection AUROCs of CFLOW-AD pipeline with the baseline ResNet18, OrthLatt-UwU-2Tap, and LS-BiorUwU-2Step encoders for hazelnut category in MVTecAD.}}
\centering
\scalebox{1.0}{
\begin{tabular}{ccc}
\hline
\multicolumn{1}{|l|}{\textbf{Models}} & \multicolumn{1}{|c|}{\textbf{SegAUROC}} & \multicolumn{1}{|c|}{\textbf{DetAUROC}}\\ \hline
\multicolumn{1}{|l|}{Baseline\cite{OrthLatt_UwU}} & \multicolumn{1}{|c|}{96.45$\%$} & \multicolumn{1}{|c|}{92.46$\%$}\\ \hline
\multicolumn{1}{|l|}{OrthLatt-UwU-2Tap\cite{OrthLatt_UwU}} & \multicolumn{1}{|c|}{97.20$\%$} & \multicolumn{1}{|c|}{89.21$\%$}\\ \hline
\multicolumn{1}{|l|}{LS-BiorUwU-2Step} & \multicolumn{1}{|c|}{97.21$\%$} & \multicolumn{1}{|c|}{92.11$\%$}\\ \hline
\end{tabular}}
\label{table:MVTecAD_hazelnut}
\end{table}
\subsection{Anomaly Detection: MVTecAD}
\begin{figure}[!t]
\begin{center}
\includegraphics[width=\linewidth]{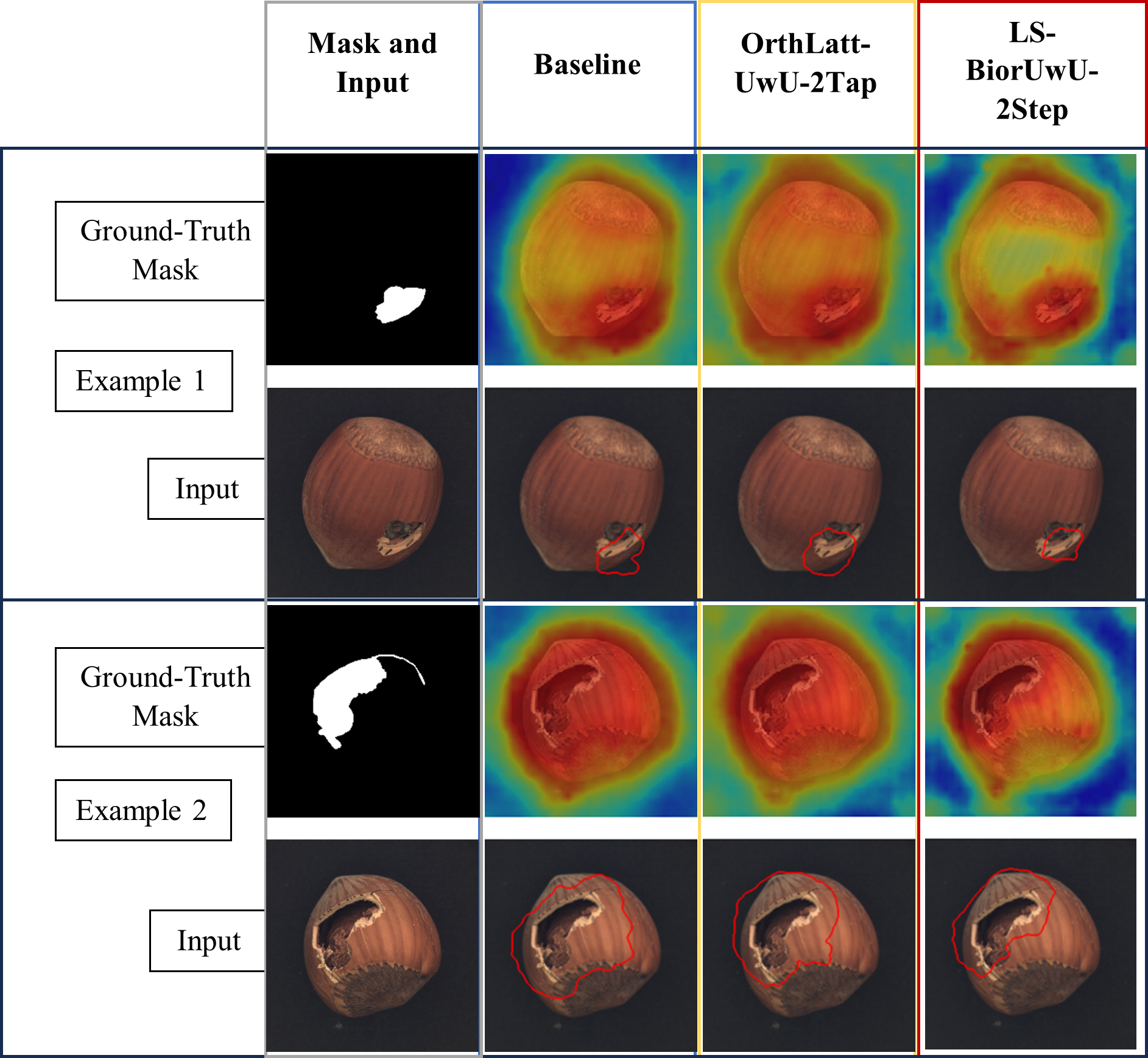}
\end{center}
\caption{\justifying{Anomaly detection on hazelnut objects from the MVTec AD dataset with two examples. From left to right: the first column presents the mask and input image, while the second to fourth columns display heatmaps and defect segmentation results from the baseline, OrthLatt-UwU-2Tap, and LS-BiorUwU-2Step, respectively.}} 
\label{fig:MVTecAD_hazelnut}
\end{figure}

In this experiment, LS-BiorUwU-2Step ResNet18, trained on the DTD dataset, was used as an encoder in the CFLOW-AD pipeline \cite{CFLOW_Gudovskiy_2022_WACV} for anomaly detection on hazelnut images from the MVTecAD dataset \cite{MVTecAD_1,MVTecAD_2}. The LS-BiorUwU-2Step ResNet18 encoder was compared against baseline ResNet18 and OrthLatt-UwU-2Tap, as shown in Table \ref{table:MVTecAD_hazelnut}. While OrthLatt-UwU-2Tap achieved a segmentation AUROC of 97.20\%, LS-BiorUwU-2Step ResNet18 slightly outperformed it with 97.21\%, while also achieving a significantly higher detection AUROC of 92.11\%. Fig. \ref{fig:MVTecAD_hazelnut} presents heatmaps from detection models, demonstrating that LS-BiorUwU-2Step ResNet18 provides better segmentation and localization performance compared to the baseline and OrthLatt-UwU-2Tap encoders.
\section{Conclusion}
\label{sec:conclude}
This study introduces a biorthogonal tunable wavelet unit based on a lifting scheme, which removes constraints on orthogonality and filter length, enabling greater design flexibility and enhancing image classification and anomaly detection in CNNs. Results show that increasing lifting steps and the order of the high-pass filter improves performance on high-frequency feature images while maintaining competitive results on low-resolution datasets. Additionally, the proposed approach balances segmentation and detection, leading to improved anomaly detection performance. However, the study has some limitations, as the current approach applies lifting steps in a single direction, from the low-pass filter to the high-pass filter, to increase the high-pass filter order. Future work can explore a dual lifting scheme to adjust both low-pass and high-pass filter orders, offering greater design flexibility. Additionally, while the current tunable biorthogonal wavelet units demonstrate good performance, their stopband attenuation can be further improved. This can be addressed by incorporating a stopband-attenuation constraint for better frequency selectivity in future research.

\bibliographystyle{ieeetr}
\bibliography{ref}

\end{document}